\documentclass[letterpaper]{article} 
\usepackage{aaai2026}  
\usepackage{times}  
\usepackage{helvet}  
\usepackage{courier}  
\usepackage[hyphens]{url}  
\usepackage{graphicx} 
\usepackage{natbib}  
\usepackage{caption} 
\usepackage{algorithm}
\usepackage{algpseudocode}
\usepackage{placeins}

\usepackage{newfloat}
\usepackage{listings}
\DeclareCaptionStyle{ruled}{labelfont=normalfont,labelsep=colon,strut=off} 
\floatstyle{ruled}
\newfloat{listing}{tb}{lst}{}
\floatname{listing}{Listing}
\usepackage{amsmath}

\usepackage{booktabs}   
\usepackage{multirow}   
\usepackage{array}      
\usepackage{makecell}   

\usepackage[table]{xcolor}          
\definecolor{lightgray}{gray}{0.9}  
\newcolumntype{g}{>{\columncolor{lightgray}}c}  

\usepackage{subcaption}  
\usepackage{caption}

\usepackage{enumitem}
\usepackage{makecell}
\usepackage{threeparttable}
\usepackage{tablefootnote}

\usepackage{arydshln}
\usepackage{amssymb}
\title{SBVR: Summation of BitVector Representation \\ for Efficient LLM Quantization}
\author{
    Wonjun Bang,
    Jongseok Park,
    Hongseung Yu,
    Kyungmin Bin\thanks{This work was done during the PhD program at Seoul National University},
    Kyunghan Lee\textsuperscript{\textdagger}
}
\affiliations{
    \textsuperscript{\rm 1,3,\textdagger}Seoul National University \\
    \textsuperscript{\rm 2}University of California, Berkeley \\
    \textsuperscript{\rm 4} Samsung SDS \\
    wjbang@snu.ac.kr \ js\_park@berkeley.edu \ hongseung@snu.ac.kr \\
    kyungmin.bin@samsung.com \ kyunghanlee@snu.ac.kr \\
}

\usepackage{bibentry}

\begin{document}
\nocopyright
\maketitle

\begin{abstract}
With the advent of large language models (LLMs), numerous Post-Training Quantization (PTQ) strategies have been proposed to alleviate deployment barriers created by their enormous parameter counts.
Quantization achieves compression by limiting the number of representable points in the data. Therefore, the key to achieving efficient quantization is selecting the optimal combination of representation points, or codes, for the given data. Existing PTQ solutions adopt two major approaches to this problem: Round-To-Nearest (RTN)-based methods and codebook-based methods.
RTN-based methods map LLM weights onto uniformly distributed integer grids, failing to account for the Gaussian-like weight distribution of LLM weights. 
Codebook-based methods mitigate this issue by constructing distribution-aware codebooks; however, they suffer from random and strided memory access patterns, resulting in degraded inference speed that is exacerbated by the limited size of GPU L1 cache.
To overcome these limitations, we propose a novel LLM quantization method, SBVR (Summation of BitVector Representation), that enables Gaussian-like code representation in a hardware-friendly manner for fast inference.
SBVR maps weight values to non-uniform representation points whose distribution follows the actual distribution of LLM weights, enabling more accurate compression.
Additionally, we design a custom CUDA kernel that allows matrix-vector multiplication directly in the SBVR format without decompression, thereby enabling high-performance execution of SBVR-compressed models.
Our evaluations of SBVR on various models demonstrate state-of-the-art perplexity and accuracy benchmark performance while delivering a 2.21×–3.04× end-to-end token-generation speedup over naive FP16 models in the 4-bit quantization regime.

\end{abstract}


\section{1\hspace{0.5cm}Introduction}
Large language models (LLMs)~\cite{achiam2023gpt4, dubey2024llama3, bai2023qwen} have demonstrated remarkable performance across a wide range of tasks, including code generation, document summarization, and even solving complex mathematical problems. Since there are strong evidences indicating that model performance is roughly proportional to model size~\cite{kaplan2020scaling, wei2022emergent}, many LLM researchers and vendors have focused on scaling up their models. However, this extreme scaling has led to substantial memory and computational resource requirements, thereby hindering easy access to LLMs. 

To mitigate this issue, myriad quantization and compression methods have been proposed to reduce the memory footprint of LLMs. Most of these techniques employ post-training quantization (PTQ), which lowers the model’s precision after it has been trained with full-precision weights. 

One large group of post-training quantization methods relies on solutions based on Round-To-Nearest (RTN) quantization. These methods map LLM weights to uniformly distributed integer points while applying various techniques that reduce quantization-induced error. Since RTN only requires scale and quantized integer weights for dequantization and further computation, it is fast and hardware-friendly. However, uniformly distributed integers do not effectively capture LLM weights, which follow a Gaussian-like distribution, thereby possessing inherent inaccuracy.

Another group of methods treats weights as sets of multi-dimensional vectors and generates specific codebooks to encode the weight vectors. Optimized codebook design can mitigate the discrepancy between the actual weight distribution and the quantization points. However, large codebook sizes and random access to codebooks for dequantization often incur significant overhead, which causes inference speed degradation.

To address these issues from previous quantization methods, we propose SBVR, a "weighted sum of bitvectors" representation that enables codebook-level weight distribution capturing capability while demonstrating RTN-like hardware efficiency for fast inference. Here, bitvectors are vectors whose elements consist of 0 or 1.

\begin{figure}[t]
  \centering
  \includegraphics[width=\linewidth]{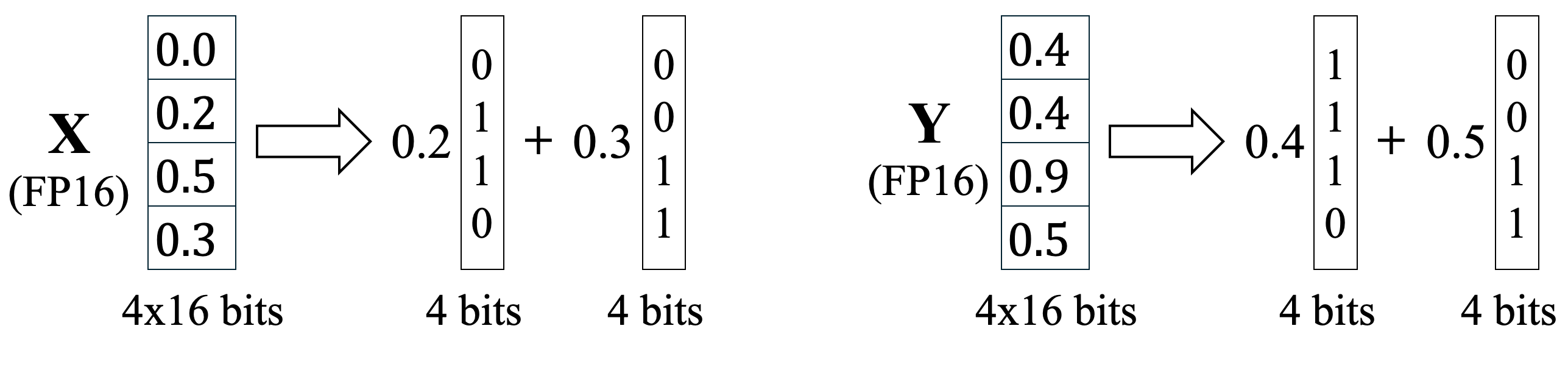}
  \caption{SBVR conversion of a given vector. Bitvectors for the weighted sum representation of a vector are obtained by identifying the set of bitvectors that minimizes the mean squared error (MSE) between the original vector and its weighted sum representation.}
  \label{fig:fig1}
\end{figure}

Given a vector and a set of coefficients $\{a_1, a_2, \dots, a_K\}$, a set of bitvectors can be selected such that the coefficient-weighted sum of these bitvectors closely approximates the vector by minimizing the Mean Squared Error (MSE).
\textbf{Figure~\ref{fig:fig1}} provides an simplified example of these bitvector assignments.  

\begin{figure}[t]
  \centering
  \begin{subfigure}[b]{0.48\columnwidth}
    \centering
    \includegraphics[width=\linewidth]{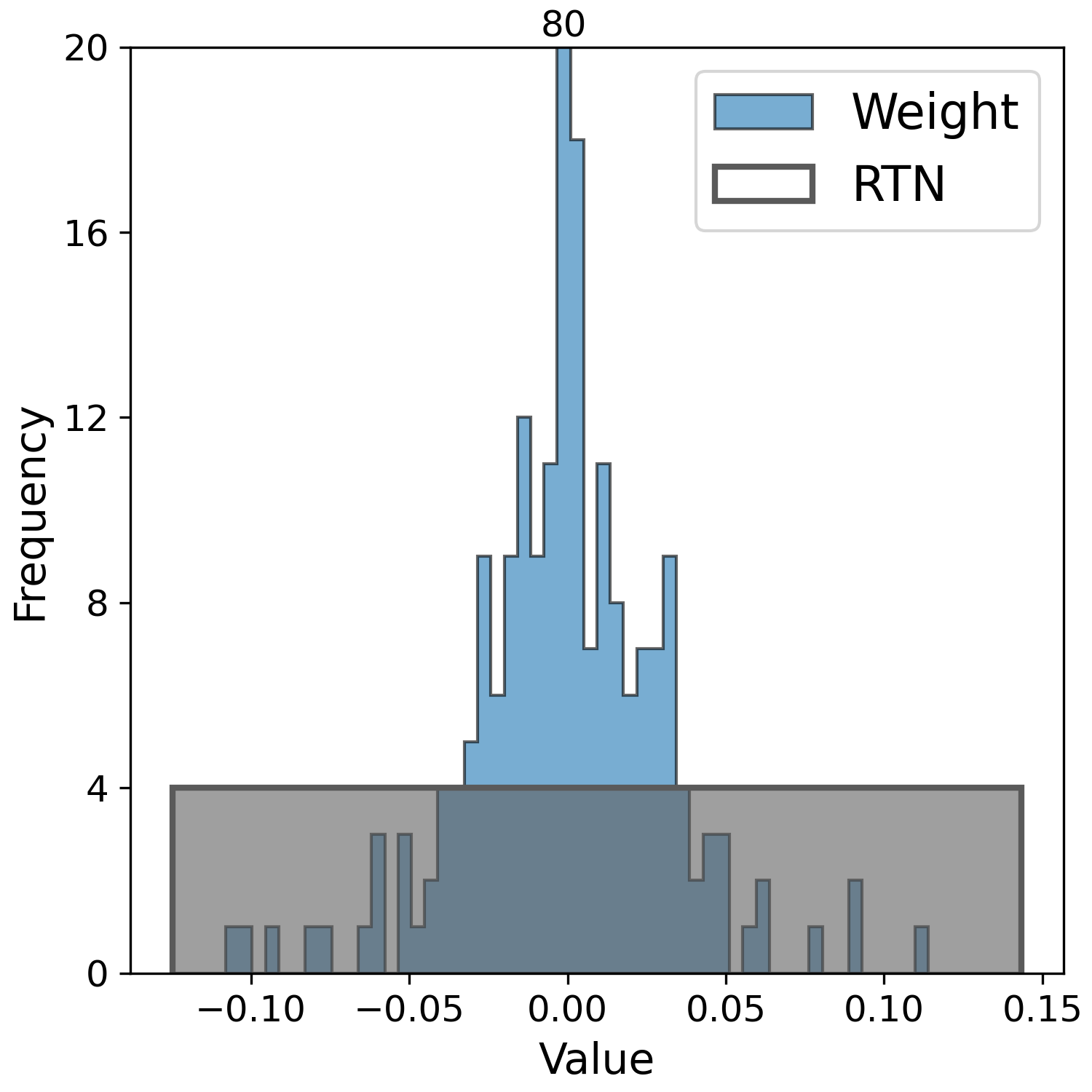}
    \caption{RTN Quantization}
    \label{fig:fig2a}
  \end{subfigure}
  \hfill
  \begin{subfigure}[b]{0.48\columnwidth}
    \centering
    \includegraphics[width=\linewidth]{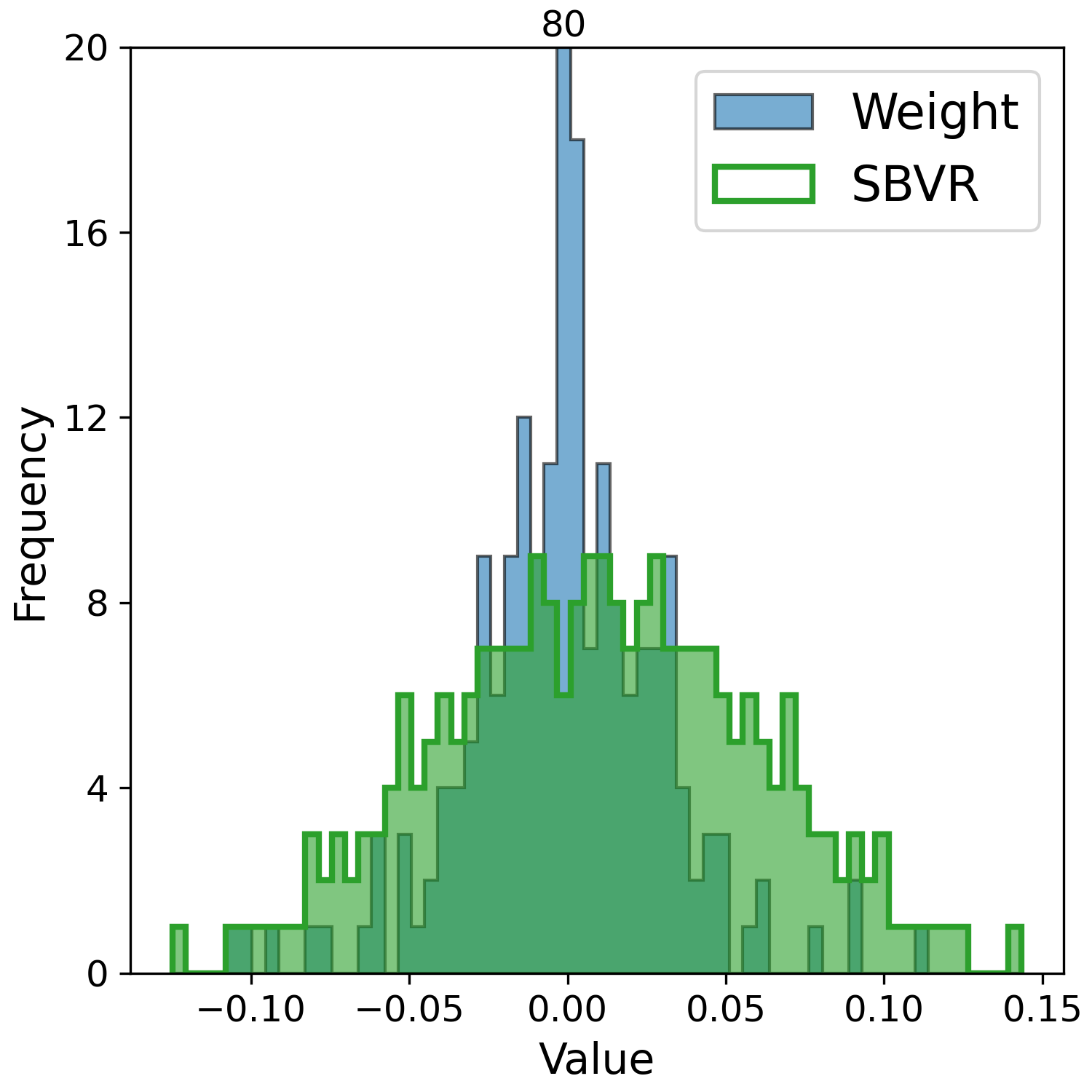}
    \caption{SBVR Quantization}
    \label{fig:fig2b}
  \end{subfigure}
  \caption{Comparison of quantization methods applied to a 256-element weight group extracted from the Llama-3.1-8B-Instruct model: (a) RTN quantization, which maps weights onto a uniform grid; (b) SBVR quantization, which effectively captures the Gaussian-like distribution of the weights.}
  \label{fig:fig2}
\end{figure}

To approximate LLM weights that follow a Gaussian-like distribution, the coefficients should be carefully determined to ensure that the subset sum of the coefficients can effectively represent the weight distribution. We empirically observed that the subset sum from a geometric series closely follows the actual weight distribution. Leveraging this observation, we designed a system that carefully selects a geometric series and applies data-dependent scaling and bias to the series, thereby allowing the scaled-and-shifted geometric series serves as effective coefficients that accurately capture the given vector distribution. 
\textbf{Figure~\ref{fig:fig2}} provides concrete support for this observation and idea.

\begin{figure}[t]
  \centering
  \includegraphics[width=\linewidth]{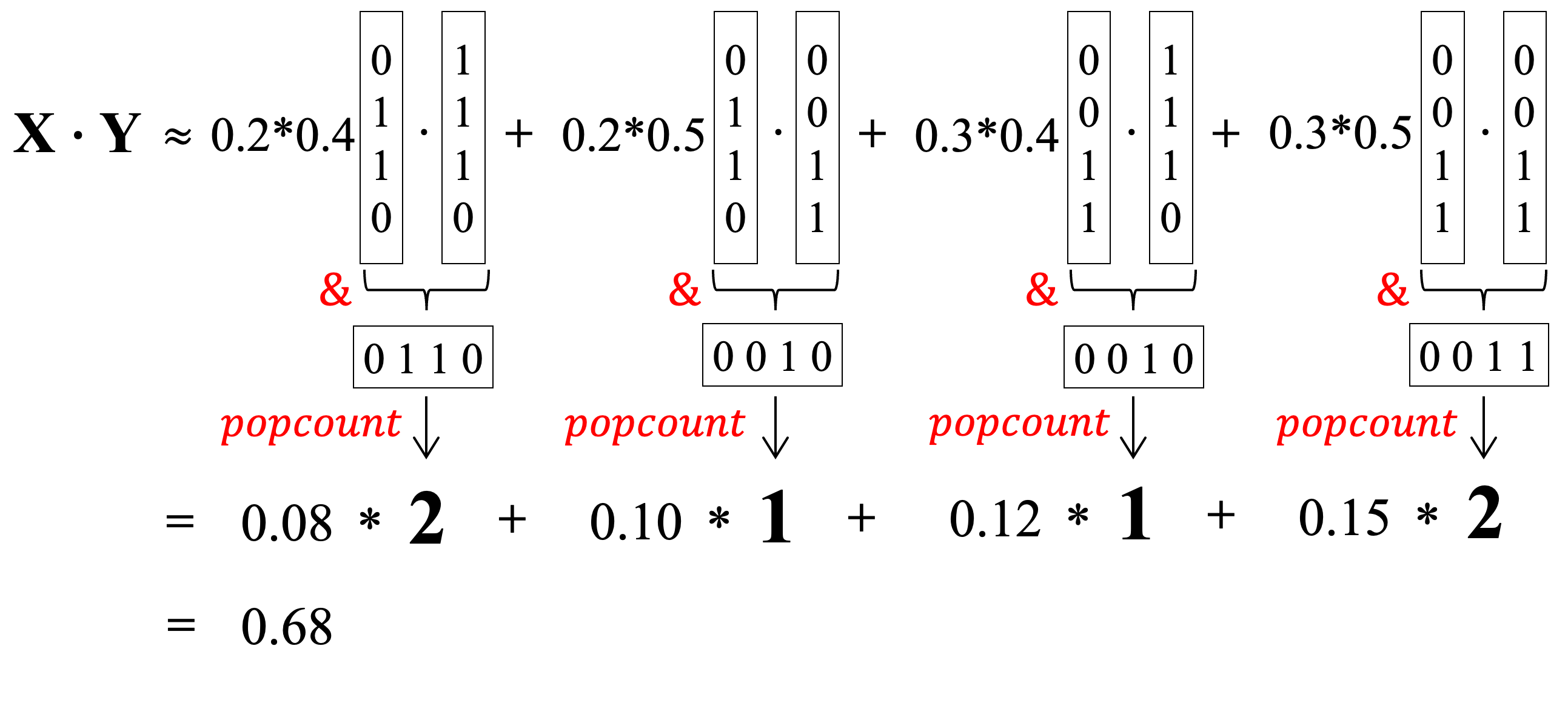}
  \caption{Computation of the inner product between two vectors represented as weighted sums of bitvectors. The inner product can be efficiently computed using a series of bitwise AND and population count operations.}
  \label{fig:fig3}
\end{figure}

The SBVR format, which represents the weighted sum of bitvectors, also provides computational benefits. By converting two floating-point vectors into the SBVR format, the inner product between them can be computed using inexpensive bitwise operations—specifically, bitwise AND and population count—instead of relying on expensive fused multiply-add (FMA) operations.
\textbf{Figure~\ref{fig:fig3}} demonstrates this process of decomposing the inner product between two vectors into bitwise AND and population count operations. Building upon this concept, we developed a GPU-friendly  General Matrix-Vector Multiplication (GEMV) kernel for SBVR-formatted weights and activations to enable faster inference.

In summary, the principal contributions of this paper are as follows:
\begin{itemize}
\item We identify inherent limitations in post-training quantization methods that must be addressed to further improve accuracy and meet stringent speed requirements.
\item We propose SBVR, a novel quantization method based on a "weighted sum of bitvectors" representation, designed to achieve both high accuracy and fast inference for LLMs while effectively overcoming the aforementioned limitations.
\item Experiments demonstrate that SBVR outperforms several prominent PTQ methods across various benchmarks, while also achieving considerable speedup in model inference. 
\end{itemize}

\section{2\hspace{0.5cm}Related Works}
We organize the related work into three subsections: conventional RTN-based element-wise quantization approaches, codebook-based quantization approaches, and other approaches that do not directly fit into these two categories.

\subsection{2.1\hspace{0.5cm}Conventional RTN-Based Quantizations}
Many prominent 3-to-4-bit targeted quantization techniques rely on RTN-based quantization, employing unique strategies to mitigate quantization errors. Outlier suppression has emerged as a primary approach for reducing quantization error in RTN-based methods. One strategy involves giving attention to outlier channels~\cite{lin2024awq}, while another approach rotates model weights or activations to mitigate outlier values~\cite{liu2024spinquant, ashkboos2024quarot}, both of which have achieved significant success. Error propagation methods for accuracy-preserving sequential quantization constitute another major branch of techniques. These methods utilize the Hessian matrix derived from inputs~\cite{frantar2022gptq} to minimize rounding errors from previous channels, or propagate quantization errors from each layer to subsequent layers~\cite{arai2025quantization} to optimize the quantization process.
All the techniques above are valid for reducing quantization errors. However, since they all rely on Round-To-Nearest quantization that maps weights onto a uniformly distributed grid, this approach is naturally suboptimal according to the rate-distortion theorem, as it encodes Gaussian-like sources using codes from a uniform distribution.

\subsection{2.2\hspace{0.5cm} Codebook-Based Quantizations}
Methods that generate well-structured codebooks to capture the weight distribution of LLMs have been proposed and have become the new mainstream approach to quantization. AQLM~\cite{egiazarian2024extreme} proposes a multiple-codebook-based additive quantization method that performs effectively at extremely low bitrates. VPTQ~\cite{liu2024vptq} further optimizes the codebook-based vector quantization method by maintaining additional codebooks for residuals and outliers. While a few methodologies attempt to mitigate the speed degradation caused by large codebook sizes~\cite{tseng2024qtip}, the majority of codebook-based approaches suffer from uncoalesced memory access problems and L1 cache misses due to exponentially growing codebooks. 

\subsection{2.3\hspace{0.5cm} Other Quantizaiton Techniques}
Several methods attempt to compress LLMs by explicitly redefining the representation of floating-point numbers. ~\cite{micikevicius2022fp8} introduce the FP8 format, which expresses floating-point numbers using 8 bits, such as E4M3 and E5M2. \cite{ocp_mx_spec_v1_2023} propose the Microscaling format (MX), which uses a per-group scale factor and employs low-precision FP formats to represent each element within a group. By adopting shared scales and asymmetric representation for each element, this approach is well-suited for quantizing LLM weights, which exhibit non-uniform distributions.

\section{3\hspace{0.5cm}Preliminaries}
As stated in Section 1, SBVR is motivated by the observation that LLM weights follow a Gaussian-like distribution and are therefore more effectively compressed with Gaussian-shaped codepoints than with the uniformly spaced codepoints used by RTN quantization. Section 3.1 revisits this observation and presents the supporting information-theoretic background. Section 3.2 introduces essential GPU architectural concepts, offers guidelines for designing hardware-friendly kernels, and explains why codebook-based quantization methods become inefficient on GPUs due to non-consecutive memory access and related overheads.

\subsection{3.1\hspace{0.5cm} Distribution of LLM Model Weights}
Numerous studies have shown that the weight data of large language models (LLMs) closely follow a Gaussian distribution~\cite{tseng2024quipshp, tseng2024qtip, si2025unveiling}. Fortunately, lossy compression of a data source following a Gaussian distribution is a classic topic of information theory, where the rate-distortion theory~\cite{shannon1959coding} provides a theoretical lower bound to the achievable data rate under a fixed distortion when compressing a Gaussian source. 

Given a Gaussian source $X\sim\mathcal{N}(\mu, \sigma^2)$, data rate of $R(D)$, and distortion of $D$ from quantizer $\hat{X}$ measured by squared error, it is proven from the rate distortion theory that 
\begin{gather}
R(D) \geq \dfrac{1}{2}log_{2}\dfrac{\sigma^2}{D},
\end{gather}
where the equality holds when $\hat{X}\sim\mathcal{N}(\mu, \sigma^2 - D)$. 

From this result, we can also observe that given a fixed data rate budget, or \textit{bitrate}, of $R$, the lowest distortion $D$ would be achieved when the quantizer closely follows a Gaussian distribution with a variance slightly below that of the source data distribution.
However, conventional integer round-to-nearest (RTN) quantization is ill-suited to achieve this. In b-bit RTN quantization, the weight elements are partitioned into groups, where each group is scaled and biased to a predefined range of integers, and each element of the group is mapped to its nearest integer using the following equation:
\begin{gather}
\Delta = \dfrac{2^{b-1}-1}{\displaystyle\max_{i}\!\left|w_i\right|} \\[4pt]
q_{i} = \operatorname{clip}\!\Bigl(
          \operatorname{round}\!\bigl(\tfrac{w_{i}}{\Delta}\bigr),
          \,-(2^{b-1}-1),\, 2^{b-1}-1
       \Bigr)
\end{gather}

Here, $\operatorname{clip}(x, a, b) := \max\!\bigl(\min(x, b), a\bigr)$.
While the biasing and scaling of conventional RTN quantization allow close reconstruction of the mean and the variance of the LLM weight data, rounding to a nearest integer in a fixed range fails to leverage the Gaussian distribution of the LLM weights, as it yields a uniformly distributed quantizer, which is proven to be suboptimal when compressing a Gaussian source.

\subsection{3.2\hspace{0.5cm} GPU-friendly Kernel}
A GPU-friendly kernel is essential for efficient execution of quantized LLMs. GPUs consist of numerous Streaming Multiprocessors (SMs), each capable of massively parallel computation across thousands of threads. GPUs feature a hierarchical memory structure with L1 and L2 caches, as well as global memory (DRAM), where access speed decreases from L1 cache to global memory. GPUs efficiently handle memory transactions through coalescing, accessing consecutive memory regions within thread groups (warps), thereby reducing transaction counts. Therefore, data layout that arranges related data consecutively in memory and kernel designs that leverage this memory layout are crucial for performance.

RTN-based quantization supports this requirement by storing scalar scales, biases, and quantized weights consecutively in memory, which can be computed  with minimal memory overhead. Conversely, codebook-based quantization methods face inherent inefficiencies due to their non-consecutive memory accesses. Moreover, their codebooks scale exponentially with target bitwidth ($2^L$), often exceeding GPU's L1 cache capacity, leading to increased cache misses, extensive indexing overhead, codebook lookups, and reconstruction processes, which significantly degrade overall inference performance.

\section{4\hspace{0.5cm} SBVR Design}
This section describes our distribution-aware efficient LLM quantization algorithm, \textbf{SBVR} (Summation of BitVector Representation), which guarantees fast and accurate inference.
   
\subsection{4.1\hspace{0.5cm} Design Overview}
To allow hardware-efficient encoding of Gaussian LLM weights, SBVR is composed of two main components: the bit vectors and a set of coefficients shared by the bit vectors. Using the weighted sum of the coefficients and the bit vectors, which we denote as SBVR, we can efficiently implement tuning of the representation points while simultaneously enabling each data element to select the representation point with minimal error. For example, given a coefficient set $[a, b, c]$, our representation points become $[0, a, b, c, a+b, a+c, b+c, a+b+c]$, and each element of the vector can freely select one of these points by assigning either 0 or 1 to the three bit vector entries weighted by the coefficients. This yields a 3-bit\footnote{The memory overhead of the coefficient set is easily amortized by the large vector size.} SBVR quantization, where we can arbitrarily select the distribution that best suits our objectives by adjusting $a$, $b$, and $c$. Although finding the optimal coefficient values for arbitrary data is an NP-hard problem, we observe that Gaussian-distributed LLM weights can be efficiently represented by setting the coefficients to a scaled-and-shifted version of a geometric series, as described in Section~4.2.

In terms of accuracy, we find it beneficial to group multiple weight elements, typically 128, into separate vectors and apply SBVR individually, similar to the grouped bias and scaling approaches used in existing RTN methods. This grouping facilitates parallelism during the encoding step and enhances accuracy by allowing optimization of the coefficient set for each vector. While the weight elements can be grouped in any orientation, we group weights along the GEMV inner-product dimension to optimize inner products within our fast inference kernel. Similarly, activations must also be converted into SBVR format to utilize our inference kernel efficiently. Thus, we propose two types of SBVR quantization algorithms: a slower but highly bit-efficient algorithm for offline weight encoding, and a lightweight, less bit-efficient algorithm for online activation encoding, as detailed in Sections~4.2 and ~4.3.

The SBVR-formatted weights and activations are sent to a GPU-friendly, high-performance General Matrix-Vector Multiplication (GEMV) kernel. The kernel eliminates the need for explicit dequantization of weights and activations and replaces costly Fused Multiply-Add (FMA) operations with relatively inexpensive bitwise operations, as elaborated in Section~4.4. Additionally, we apply complementary techniques from other research alongside SBVR quantization to further enhance its quantization performance. Further details are provided in Section~4.5.

\begin{figure}[t]        
  \centering
  \includegraphics[width=\columnwidth]{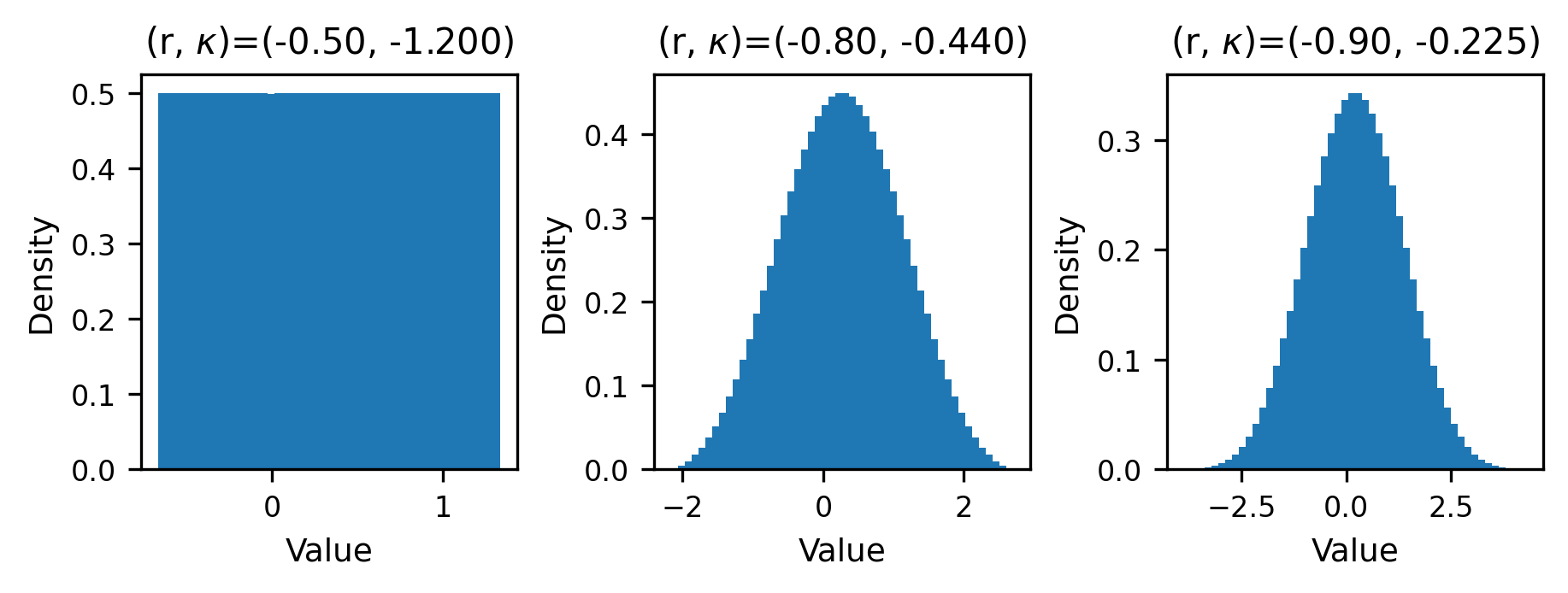}
  \caption{Generated representation point distributions for three distinct r values. The s and b values are set to 1 and 0, respectively, for demonstration purposes. Excess kurtosis $\kappa$ represents the "tail thickness" of a distribution. The Gaussian distribution has zero excess kurtosis, while the uniform distribution has an excess kurtosis of -1.2. By changing the r value from -1 to -0.5, we can tune the excess kurtosis to better fit the representation points to the data distribution.}
  \label{fig:fig_dist}
\end{figure}





\subsection{4.2\hspace{0.5cm} Offline Weight Quantization}
We employ three variables---ratio (r), scale (s), and bias (b)---to generate a search space for our coefficient set. Each variable changes a different characteristic in the distribution of the representable points: the ratio changes the shape, the scale changes the variance, and the bias change the mean of the distribution. By adjusting each variable, we can fit our representable point distribution to that of the weight data. 
We denote the number of coefficients in the coefficient set as $K$, which in turn determines the target bitrate of the SBVR quantization.
\begin{equation}
\begin{aligned}
  \left\{\,s_j r_i^{0} + b_k,\;
           s_j r_i^{1} + b_k,\;
           \cdots,\;
           s_j r_i^{K-1} + b_k \,\right\}
  \\[4pt] 
  1 \le i \le \lvert\it{R}\rvert,\quad
  1 \le j \le \lvert\it{S}\rvert,\quad
  1 \le k \le \lvert\it{B}\rvert
\end{aligned}
\label{eq:search-space}
\end{equation}

Eq~\ref{eq:search-space} shows the coefficient set generated using a scaled-and-shifted geometric series with variables $r_i$, $s_j$, and $b_k$ from the set of cadidate variables $R, S$ and $B$, repectively. The size of the SBVR parameter search space is therefore $|R|\times|S|\times|B|$.
\textbf{Figure~\ref{fig:fig_dist}} illustrates the effect of changing r on the generated representation points. As illustrated in the figure, entries in the search space can generate various forms of distributions, ranging from Uniform to Gaussian-like distributions. 
As such, SBVR is strictly a superset of RTN quantization and can handle many variants of Gaussian-like distributions in addition, enabling accurate quantization of the Gaussian-like LLM weights.

As stated in Eq~\ref{eq:search-space}, $R, S$ and $B$ determine the overall landscape of the search space. 
Therefore, it is crucial to judiciously determine the candidate sets $R, S,$ and $B$ for the parameters $\left(r, s, b\right)$. 
We used the following equation 5 to 11 to define the candidate sets for the parameters.
\newcommand{\Nratio}{N_{\!\mathrm{ratio}}}
\newcommand{\Nbias}{N_{\!\mathrm{bias}}}
\newcommand{\Nscale}{N_{\!\mathrm{scale}}}


\begin{gather}
  \it{R} =  \{\, r_i \mid 0 \le i < \Nratio \} \\[4pt]
  \it{S} = \{\, s_{\min} + (j+1)s_{\text{gran}} \mid 0 \le j < \Nscale \} \\[4pt]
  \it{B} = \{\, b_{\min} + k\times b_{\text{gran}} \mid 0 \le k < \Nbias \} \\[4pt]
  s_{\max} = 1.1\bigl(\max(D) - \min(D)\bigr),\\[4pt]
  b_{\max} = {2\,\text{abs}(\text{avg}(D))}/{K}\\[4pt]
  s_{\min} = 2\,q_{95}(D),\quad
  b_{\min} = -b_{\max}\\[4pt]
  s_{\text{gran}} = \dfrac{s_{\max}-s_{\min}}{\Nscale}, \quad
  b_{\text{gran}} = \dfrac{b_{\max}-b_{\min}}{\Nbias}
\end{gather}

\noindent Here, $\Nratio$, $\Nscale$, and $\Nbias$ are the size of each search dimension that can be tuned, and $D$ is the target weight data. $R$ consists of a superset of two disjoint sets, where each set is generated using the same procedure employed to generate $S$ and $B$.
$q_{95}(D)$ denotes the 95th percentile of the given data.
Further details, such as the actual values used for the experiments, can be found in the Technical Appendix. 

\begin{algorithm}[t]
  \caption{SBVR Encoding}
  \label{alg:sbvr}

  \textbf{Input}: weight vector $X$, search space $\mathbb{S}$\\
  \textbf{Output}: entry $e^\star$ minimizing mean-squared error (MSE)\\

  \begin{algorithmic}[1]
    \State $e^\star \gets \text{nil}$,\; $\text{bestMSE} \gets \infty$
    \ForAll{$e \in \mathbb{S}$}
      \State $\text{sums} \gets \textsc{AllSubsetSums}(e)$
      \State $\text{errorSum} \gets 0$
      \ForAll{$x \in X$}
        \State $n \gets \textsc{Nearest}(\text{sums}, x)$
        \State $\text{errorSum} \gets \text{errorSum} + (x - n)^2$
      \EndFor
      \State $\text{mse} \gets \text{errorSum} / |X|$
      \If{$\text{mse} < \text{bestMSE}$}
        \State $e^\star \gets e$,\; $\text{bestMSE} \gets \text{mse}$
      \EndIf
    \EndFor
    \State \textbf{return} $e^\star$
  \end{algorithmic}

  \begin{flushleft}
    \footnotesize
    \textbf{Notes} \\
    (1) \textsc{AllSubsetSums}: Retrieves all subset sums from entry e\\
    (2) \textsc{Nearest}: Return the sum that is closest to $x$
  \end{flushleft}
\end{algorithm}

\textbf{Algorithm~\ref{alg:sbvr}} illustrates the process of determining the optimal entry from the search space for each weight vector. After selecting the entry that minimizes the mean squared error (MSE) for the given weight vector, the coefficients and bitvectors can be determined. Each element in the selected entry becomes a coefficient. For each element $x_{i}$ in the weight vector, we identify the subset of elements from the entry whose subset sum is closest to $x_{i}$. We then mark the selected elements as 1 and the others as 0, forming a binary array. When using K-bit SBVR (SBVR-K), this K-dimensional binary array represents the array of the $i^{\text{th}}$ elements of the bitvectors.

We observe that many vectors in the SBVR format share similar coefficient sets. To accelerate the search process, we maintain a cache of previously selected variables $r$, $s$, and $b$. The search algorithm first checks the cache to determine if any previously selected combinations of $r$, $s$, and $b$ provide sufficient accuracy\footnote{We maintain a moving average of the quantization error. If the error from the cached values is below this moving average, the cached values are used.} before exploring the entire search space. In many cases, the cache hit rate reaches up to 90\%, greatly reducing quantization time and significantly decreasing the size of the coefficient cache.

\subsection{4.3\hspace{0.5cm} Online Activation Conversion on Decoding Stage}
Unlike LLM weights, which are converted to the SBVR format offline, activations must be converted online. Consequently, the search-space-based method employed for weight conversion is computationally infeasible for activations due to its prohibitive overhead. Therefore, we adopt the conventional integer-grid method for activation conversion, which, while suboptimal in terms of accuracy, ensures significantly lower computational costs compared to exhaustive search-space traversal. First, we partition the activations into a set of vectors and predetermine their SBVR-format coefficients as power-of-two values. Then, for each activation vector, we compute a per-vector scale factor by dividing the absolute maximum value of the vector by $2^l$, where $l$ denotes the target bitwidth for the activations. Finally, during the actual conversion process, we multiply this per-vector scale factor by the power-of-two coefficients.

Equation~\ref{eq:activation-coeffs} illustrates the final coefficients obtained after this conversion, with $s$ representing the per-vector scale factor and the bias set to 0. We set $l$ to 8 for near-lossless quantization of activations~\cite{10.5555/3600270.3602468}, minimizing the accuracy degradation inherent in integer-grid RTN quantization. The corresponding bitvectors are then obtained using the same procedure as in weight quantization.

\begin{gather}
\bigl\{ -2^{\,{(b-1)}}s,\; s,\; 2s,\; \cdots,\; 2^{\,{(b-2)}}s \bigr\}
\label{eq:activation-coeffs}
\end{gather}

\subsection{4.4\hspace{0.5cm} GPU-Friendly Kernel Design of SBVR}
The goal of SBVR is not only to achieve accurate quantization of LLM weights but also to enable fast LLM inference during the decoding stage. To achieve high inference speed, we designed a GPU-friendly GEMV kernel. After converting weights and activations to SBVR, we separately store coefficients and bitvectors. Specifically, the bitvectors are partitioned into 32-bit integers and stored as a tensor in a contiguous memory region. For coefficients, we utilize a coefficient cache containing all coefficient sets required for decoding, as well as a coefficient index that identifies the specific coefficient set used by each K-bit bitvector set.
For activations, we maintain predefined power-of-two coefficients in the GPU's constant memory, storing only the bitvectors and per-vector scale factors. At runtime, these scale factors are multiplied by the predefined coefficients to reconstruct the final coefficients.

During GEMV—the inner product between an activation vector and each weight row—the kernel first fetches the relevant bitvectors, coefficients, and scales, loading the SBVR-formatted weights and activations into registers. Then, the kernel multiplies corresponding coefficients and performs the inner product between the bitvectors of weights and activations. The inner products between bitvectors can be decomposed into two bitwise operations: bitwise AND and population count. After performing the AND operation between bitvectors, we apply the population count, which counts the number of set bits (1's) resulting from the bitwise AND operation performed immediately prior. Finally, we accumulate all multiplication results of the coefficients and population counts into a single FP16 element. For clarity, readers are encouraged to revisit \textbf{Figure~\ref{fig:fig3}}, which provides a comprehensive example illustrating the entire computation process.

This design offers two key benefits. First, it eliminates the need for per-element de-quantization steps during computation. Second, when employing K-bit and N-bit SBVR quantization for weights and activations, respectively, only $KN$ Fused Multiply-Add (FMA) operations are required per inner product between these two vectors. For instance, in the W4A8 configuration with 128-dimensional vectors, this approach reduces the number of FMA operations by a factor of $4\times8/128=4$. Consequently, this kernel design not only reduces memory traffic by loading only uint32-packed bitvectors but also decreases computationally expensive FMA operations by replacing them with efficient bitwise operations.

\subsection{4.5\hspace{0.5cm} Additional Techniques Applied to SBVR} 
Since SBVR does not have inherent dependencies on other existing techniques, methods such as outlier suppression and error propagation can be combined orthogonally with SBVR to further enhance quantization accuracy. 
To ensure that LLM weights closely follow a Gaussian-like distribution and to mitigate the negative impact of outliers on quantization performance, we applied random Hadamard rotation merging~\cite{liu2024spinquant} to the weights prior to quantization. 
Additionally, we employed a coarse-grained variant of GPTQ~\cite{frantar2022gptq}, which propagates quantization errors across multiple columns simultaneously, thereby further reducing quantization-induced errors.

\newcommand{\mypara}[1]{\noindent\textbf{#1.}\;}
\renewcommand{\mypara}[1]{%
  \par\medskip          
  \noindent\textbf{#1.}\;
}
\section{5\hspace{0.5cm}Experiment}
In this section, we introduce the experimental setup, including the models we used, evaluation metrics, baseline methods, test environment, and implementation details for the experiments. We also compare the benchmark results of our methodology, SBVR, with other baseline quantization methods.

\subsection{5.1\hspace{0.5cm} Experiment Setting}
\mypara{Models} We test our methodology on various models with different sizes. We used Llama3~\cite{dubey2024llama3} series models: Llama-3.2-1B-Instruct, Llama-3.2-3B-Instruct, Llama-3.1-8B-Instruct, and Llama-3.1-70B-Instruct. We also used the latest Qwen~\cite{bai2023qwen} models: Qwen3-0.6B, Qwen3-1.7B, Qwen3-8B, and Qwen3-14B. Finally, we tested the DeepSeek-R1~\cite{guo2025deepseek} distilled model, DeepSeek-R1-Distil-Qwen-7B.

\mypara{Evaluation Metrics} To quantify the accuracy of the quantized model, we measure perplexity (PPL) on WikiText-2~\cite{merity2016pointer} and evaluate performance by measuring zero-shot accuracy on the following six commonsense benchmarks for the models listed in Section~5.1: ARC-Easy, ARC-Challange~\cite{clark2018think}, CommonsenseQA~\cite{talmor2018commonsenseqa}, HellaSwag~\cite{zellers2019hellaswag}, PIQA~\cite{bisk2020piqa}, and WinoGrande~\cite{sakaguchi2021winogrande}.
To measure inference speed during the decoding stage, we evaluated the Time-Per-Output-Token (TPOT) of the models when quantization was applied. 

\mypara{Baseline Methods} We compare our proposed method, SBVR, against several popular and state-of-the-art baseline quantization methods in the 4-bit regime, including GPTQ~\cite{frantar2022gptq}, AWQ~\cite{lin2024awq}, SpinQuant~\cite{liu2024spinquant}, and VPTQ~\cite{liu2024vptq}.
For inference speed evaluation, we use RTN-based methods equipped with custom kernels optimized for speed (GPTQ, AWQ), as well as the codebook-based method VPTQ as baselines.
To evaluate the effectiveness of the SBVR kernel against other W4A8 methods, we also include QQQ~\cite{zhang2024qqq} in our speed comparison experiments.
Note that due to the limited availability of publicly released VPTQ models on Hugging Face~\cite{wolf2019huggingface}, some benchmarks could not be measured for this method.

\mypara{Test Environment \& Implementation Details} All speed measurements are conducted on a single machine equipped with NVIDIA GeForce RTX 3090 GPUs. The SBVR encoding algorithm is implemented using PyTorch. Customized kernels for GEMV operations and online activation conversion are implemented in C++ and CUDA, and are imported into PyTorch as an extension.
For the decoding stage, we fix the SBVR configuration to W4A8, with SBVR-4 for weights and SBVR-8 for activations.
Unlike the decoding stage, we can leverage tensor cores to accelerate massive General Matrix Multiplication (GEMM) operations, amortizing the dequantization overhead across multiple tokens. Therefore, we design a prefill kernel that decompresses SBVR weights into FP16 and transfers the recovered weight segments to tensor cores for GEMM computation. 


\subsection{5.2\hspace{0.5cm} Experimental Results}
This section presents the evaluation results for the accuracy and inference speed of our quantization method. 

\newcommand{\tabscale}{0.87} 

\newcommand{\doublemidrule}[1]{%
  \specialrule{\lightrulewidth}{0pt}{#1}%
  \specialrule{\lightrulewidth}{#1}{0pt}%
}
\begin{table}[t]
  \centering
  \scriptsize
  \setlength{\tabcolsep}{3pt}
  \setlength{\doublerulesep}{0.8pt} 
  \renewcommand{\arraystretch}{1.2}
  \scalebox{\tabscale}{%
    \begin{tabular}{@{}l l c:c:c c c c c c@{}}
      \toprule
      Model & Method & 
      \multicolumn{1}{c}{\cellcolor{white}{Bits}} & 
      \multicolumn{1}{c}{\cellcolor{white}{PPL\,$\downarrow$}} &
      {CQA\,$\uparrow$} & {ARC-C\,$\uparrow$} & {ARC-E\,$\uparrow$} 
      & {HS\,$\uparrow$} & {PIQA\,$\uparrow$} & {WG\,$\uparrow$} \\
      \hline\hline
      \multirow[c]{6}{*}{\makecell[l]{3.2-1B\\\scriptsize Instruct}}
        & FP16       & 16 & 13.72 & 55.36 & 35.84 & 68.52 & 45.21 & 74.37 & 59.67 \\
        & AWQ        &  4 & 15.05 & 52.33 & 34.98 & 66.96 & 44.07 & 72.63 & 59.19 \\
        & GPTQ       &  4 & 90.11 & 20.23 & 23.38 & 51.68 & 31.18 & 63.55 & 55.09 \\
        & SpinQuant  &  4 & 14.58 & 49.14 & 34.81 & 67.34 & 43.75 & 72.69 & 59.35 \\
        & Ablation  &  4 & 14.58 & 49.14 & 34.04 & 67.17 & 43.71 & 72.52 & 58.56 \\
        \rowcolor{gray!15}
        \cellcolor{white}
        & SBVR       &  4 & 14.41 & 51.84 & 34.22 & 66.96 & 44.30 & 72.69 & 59.91 \\
      \hline
      \multirow[c]{6}{*}{\makecell[l]{3.2-3B\\\scriptsize Instruct}}
        & FP16       & 16 &  11.45 & 67.90 & 43.69 & 74.20 & 52.24 & 75.73 & 67.32 \\
        & AWQ        &  4 &  12.15 & 65.19 & 41.55 & 73.02 & 51.60 & 74.86 & 67.48 \\
        & GPTQ       &  4 &  18.97 & 60.36 & 39.08 & 70.88 & 49.95 & 72.36 & 61.88 \\
        & SpinQuant  &  4 &  11.65 & 65.03 & 41.38 & 73.36 & 51.43 & 74.27 & 67.48 \\
        & Ablation  &  4 &  12.49 & 67.40 & 41.47 & 72.73 & 51.01 & 73.83 & 67.40 \\
        \rowcolor{gray!15}
        \cellcolor{white}
        & SBVR       &  4 &  12.10 & 65.85 & 40.96 & 73.23 & 51.29 & 74.05 & 67.17 \\
      \hline
      \multirow[c]{6}{*}{\makecell[l]{3.1-8B\\\scriptsize Instruct}}
        & FP16       & 16 &  7.50 & 77.23 & 51.79 & 81.82 & 59.11 & 79.92 & 73.88 \\
        & AWQ        &  4 &  7.85 & 76.74 & 51.19 & 80.93 & 58.43 & 79.21 & 73.72 \\
        & GPTQ       &  4 &  7.91 & 71.58 & 48.72 & 78.70 & 56.88 & 78.84 & 71.67 \\
        & SpinQuant  &  4 &  7.76 & 75.92 & 51.45 & 80.72 & 57.90 & 79.49 & 73.95 \\
        & VPTQ       &  4 &  7.70 & 76.66 & 51.45 & 81.23 & 58.68 & 80.20 & 73.40 \\
        & Ablation  &  4 &  7.81 & 75.68 & 50.26 & 80.30 & 58.33 & 78.84 & 73.40 \\
        \rowcolor{gray!15}
        \cellcolor{white}
        & SBVR       &  4 &  7.71 & 76.00 & 51.88 & 81.44 & 58.27 & 79.49 & 73.32 \\
      \hline
      \multirow[c]{4}{*}{\makecell[l]{3.1-70B\\\scriptsize Instruct}}
        & FP16       & 16 &  4.40 & 80.59 & 62.46 & 86.74 & 65.18 & 82.97 & 79.48 \\
        & AWQ        &  4 &  4.82 & 80.84 & 62.12 & 86.57 & 64.89 & 83.41 & 79.32 \\
        & GPTQ       &  4 &  4.77 & 79.93 & 61.18 & 86.07 & 64.65 & 82.86 & 77.98 \\
        & VPTQ       &  4 &  4.65 & 80.02 & 61.01 & 86.32 & 64.92 & 82.86 & 78.77 \\
        \rowcolor{gray!15}
        \cellcolor{white}
        & SBVR       &  4 &  4.71 & 79.85 & 62.63 & 86.53 & 64.49 & 82.92 & 80.27 \\
      \bottomrule
    \end{tabular}
  }
  \caption{Benchmark results for the Llama-3.1 and 3.2 Instruct model series. CQA, HS, WG represent CommonsenseQA, HellaSwag, and Winogrande, respectively. Lower perplexity values indicate better performance, while higher commonsense benchmark accuracy indicates better performance. Ablation refers to the SBVR implementation without using the Guassian representation points.}
  \label{tab:quant_bench_llama}
\end{table}

\begin{table}[t]
  \centering
  \scriptsize
  \setlength{\tabcolsep}{3pt}
  \setlength{\doublerulesep}{0.8pt} 
  \renewcommand{\arraystretch}{1.2}
  \begin{threeparttable}
  \scalebox{\tabscale}{%
    \begin{tabular}{@{}l l c:c:c c c c c c@{}}
      \toprule
      Model & Method & 
      \multicolumn{1}{c}{\cellcolor{white}{Bits}} & 
      \multicolumn{1}{c}{\cellcolor{white}{PPL\,$\downarrow$}} &
      {CQA\,$\uparrow$} & {ARC-C\,$\uparrow$} & {ARC-E\,$\uparrow$} 
      & {HS\,$\uparrow$} & {PIQA\,$\uparrow$} & {WG\,$\uparrow$} \\
        \hline\hline
      \multirow{4}{*}{0.6B}
        & FP16 & 16 & 22.24 & 46.60 & 31.40 & 60.86 & 37.54 & 67.68 & 56.20 \\
        & AWQ  &  4 & 26.04 & 36.94 & 28.75 & 51.09 & 36.27 & 66.00 & 56.12 \\
        & GPTQ &  4 & 23.64 & 35.71 & 29.27 & 57.53 & 36.11 & 64.47 & 55.01 \\
        \rowcolor{gray!15}
        \cellcolor{white}
        & SBVR &  4 & 25.09 & 40.05 & 28.75 & 56.61 & 36.09 & 65.40 & 55.09 \\
      \hline
      \multirow{4}{*}{1.7B}
        & FP16 & 16 & 17.67 & 64.21 & 39.85 & 72.18 & 46.14 & 72.47 & 61.09 \\
        & AWQ  &  4 & 19.14 & 63.72 & 38.14 & 71.30 & 44.77 & 70.24 & 61.64 \\
        & GPTQ &  4 & 18.83 & 60.28 & 35.15 & 64.90 & 43.31 & 70.40 & 59.12 \\
        \rowcolor{gray!15}
        \cellcolor{white}
        & SBVR &  4 & 18.40 & 60.28 & 41.30 & 73.40 & 44.97 & 70.78 & 60.85 \\
      \hline
      \multirow{4}{*}{8B}
        & FP16 & 16 &  10.26 & 78.71 & 55.80 & 83.50 & 57.15 & 76.82 & 67.72 \\
        & AWQ  &  4 &  10.71 & 77.97 & 54.01 & 82.41 & 56.24 & 77.04 & 67.40 \\
        & GPTQ &  4 &  10.56 & 74.94 & 51.28 & 80.81 & 55.66 & 75.52 & 67.88 \\
        \rowcolor{gray!15}
        \cellcolor{white}
        & SBVR &  4 &  10.35 & 79.03 & 54.01 & 82.53 & 56.42 & 76.55 & 67.40 \\
      \hline
      \multirow{4}{*}{14B\tnote{*}}
        & FP16 & 16 &  9.14 & 80.18 & 58.70 & 84.26 & 61.05 & 80.03 & 72.69 \\
        & AWQ  &  4 &  9.5 & 78.71 & 58.11 & 83.54 & 60.35 & 80.03 & 71.90 \\
        & GPTQ &  4 &  9.5 & 79.69 & 57.94 & 83.29 & 59.78 & 79.16 & 72.38 \\
        \rowcolor{gray!15}
        \cellcolor{white}
        & SBVR &  4 &  9.28 & 79.61 & 58.02 & 84.09 & 60.35 & 79.60 & 72.61 \\
      \hline
      \multirow{4}{*}{DS-7B\tnote{*}}
        & FP16 & 16 &  26.81 & 56.02 & 42.41 & 69.65 & 46.64 & 71.00 & 60.46 \\
        & AWQ  &  4 &  29.26 & 56.67 & 41.04 & 67.68 & 45.51 & 70.51 & 59.67 \\
        & GPTQ &  4 &  28.41 & 53.89 & 39.51 & 67.00 & 45.53 & 70.89 & 61.09 \\
        \rowcolor{gray!15}
        \cellcolor{white}
        & SBVR &  4 &  27.97 & 53.81 & 41.64 & 67.72 & 46.02 & 69.64 & 60.46 \\
      \bottomrule
    \end{tabular}}
    \end{threeparttable}
  \caption{Benchmark results for the Qwen3 model series and DeepSeek-R1-Distill-Qwen-7B. The table follows the same format as Table~\ref{tab:quant_bench_llama}. Hadamard rotation was not applied to the 14B and DS-7B models since there is no matching Hadamard matrix for their hidden dimensions.}
  \label{tab:quant_bench_qwen}
\end{table}

\mypara{Evaluation on Llama Models} Table~\ref{tab:quant_bench_llama} presents the perplexity (PPL) and commonsense benchmark results for the Llama 3 model series. For both the 1B and 8B models, SBVR outperforms RTN-based quantization methods, including AWQ, GPTQ, and SpinQuant, and achieves results comparable to the codebook-based quantization method VPTQ. Specifically, SBVR achieves lower PPL and superior performance across various commonsense benchmarks.

We further conduct an ablation study to assess the effectiveness of the SBVR coding scheme by implementing a quantization method that retains all SBVR techniques, such as Hadamard rotations and GPTQ error propagation, but replaces the Gaussian representation points with the integer grid employed by existing RTN-based methods. As shown in the table, SBVR consistently surpasses this ablation implementation in terms of both PPL and commonsense task performance, providing additional evidence of the effectiveness of our quantization approach relative to RTN-based quantization methods.

\mypara{Evaluation on Qwen Models} Table~\ref{tab:quant_bench_qwen} presents the perplexity and commonsense benchmark results for the Qwen3 model series and the DeepSeek-R1-Distill-Qwen-7B model.
Except for the 0.6B model, SBVR demonstrates the lowest PPL among the baseline methods for every model included in the table. SBVR also ranks at or near the top across diverse commonsense benchmarks.
For instance, the Qwen3-8B model achieves the highest scores across all evaluated tasks except PIQA, where it attains a nearly identical score to the best-performing AWQ method. 

\begin{table}[t]
  \centering
  \footnotesize 
  \setlength{\tabcolsep}{3pt}
  \renewcommand{\arraystretch}{1.35}

  \resizebox{\columnwidth}{!}{%
  \begin{tabular}{@{}c|c c c c c c@{}}
    \hline
    \textbf{Model/Methods} & \textbf{Naive Torch (ms)} & \textbf{SBVR (ms)} & \textbf{GPTQ (ms)} & \textbf{AWQ (ms)} & \textbf{VPTQ (ms)} & \textbf{QQQ (ms)}  \\
    \hline
    3.2-1B & 10.81 & 3.56 & 4.18 & 2.59 & x & x \\
    3.2-3B & 18.25 & 6.73 & 7.09 & 5.00 & x & x \\
    3.1-8B & 22.68 & 10.24 & 10.51 & 8.61 & 28.49 & 12.66\footnotemark[3] \\
    \hline
  \end{tabular}}
  \caption{Time-per-output-token (TPOT) comparison of various quantization methods for Llama models. Naive Torch represents Hugging Face models with FP16 torch kernels, while the other methods represent 4-bit quantized models with their own custom kernels. X denotes publicly unavailable models.}
  \label{tab:latency}
\end{table}

\footnotetext[3]{
3-8B was used instead of 3.1-8B as QQQ does not support 3.1 models. Since 3-8B and 3.1-8B have identical structures except for the minimal RoPE embedding overhead, their execution times can be considered equivalent.}

\mypara{Speed Measurements} We measure the inference speed of SBVR-quantized models and compare them with models quantized using other methods. For a fair comparison, we use the model implementation from Huggingface~\cite{wolf2019huggingface} and only replace the linear layers with layers containing customized kernels for each method~\cite{qubitium2024gptqmodel, autoawq_2025}. Additionally, we apply CUDAGraph~\cite{cuda_graphs_guide_12_9} to all quantized models to reduce excessive overhead from GPU kernel launching. We fix the batch size to 1 for all inference measurements.

Table~\ref{tab:latency} compares the time-per-output-token (TPOT) across the Llama model series.
Although AWQ achieves the best overall performance, SBVR attains comparable performance to AWQ and consistently outperforms the GPTQ kernel across all evaluated models, while delivering superior accuracy compared to RTN-based methods.
Furthermore, when compared to the codebook-based method VPTQ, SBVR demonstrates substantial acceleration, achieving up to a 2.78$\times$ speedup and surpassing the performance of the W4A8 kernel utilized by QQQ.

\section{6\hspace{0.5cm}Conclusion}

This study introduces SBVR, a novel quantization technique for large language models (LLMs) that effectively maintains quantization accuracy while achieving high inference speed. 
SBVR maps the weights of LLMs onto quantization points approximating a Gaussian distribution and introduces a specialized data format and computational kernel optimized for rapid GEMV operations on GPUs. 
Extensive experimental results confirm that SBVR accurately captures the Gaussian-like characteristics of LLM weight distributions and effectively leverages GPU architectures, resulting in significantly enhanced inference speed.
\clearpage   
\bibliography{aaai2026}

\clearpage            

\end{document}